# The Impact of Loss Functions and Scene Representations for 3D/2D Registration on Single-view Fluoroscopic X-ray Pose Estimation

Chaochao Zhou[1,*], Syed Hasib Akhter Faruqui[1], Abhinav Patel[1], Ramez N. Abdalla[1], Michael C. Hurley[1,2,3], Ali Shaibani[1,2,3], Matthew B. Potts[1,3], Babak S. Jahromi[1,3], Sameer A. Ansari[1,2,3], Donald R. Cantrell[1,2,*]

[1]Department of Radiology, [2]Department of Neurology, and [3]Department of Neurological Surgery, Northwestern University and Northwestern Medicine, Chicago, IL, United States

*Corresponding Authors:
Chaochao Zhou: chaochao.zhou@northwestern.edu
Donald R. Cantrell: donald.cantrell@nm.org

## Abstract

Many tasks performed in image-guided procedures can be cast as pose estimation problems, where specific projections are chosen to reach a target in 3D space. In this study, we first develop a differentiable projection (DiffProj) rendering framework for the efficient computation of Digitally Reconstructed Radiographs (DRRs) with automatic differentiability from either Cone-Beam Computerized Tomography (CBCT) or neural scene representations, including two newly proposed methods, Neural Tuned Tomography (NeTT) and masked Neural Radiance Fields (mNeRF). We then perform pose estimation by iterative gradient descent using various candidate loss functions, that quantify the image discrepancy of the synthesized DRR with respect to the ground-truth fluoroscopic X-ray image. Compared to alternative loss functions, the Mutual Information loss function can significantly improve pose estimation accuracy, as it can effectively prevent entrapment in local optima. Using the Mutual Information loss, a comprehensive evaluation of pose estimation performed on a tomographic X-ray dataset of 50 patients' skulls shows that utilizing either discretized (CBCT) or neural (NeTT/mNeRF) scene representations in DiffProj leads to comparable performance in DRR appearance and pose estimation (3D angle errors: mean ≤ 3.2° and 90% quantile ≤ 3.4°), despite the latter often incurring considerable training expenses and time. These findings could be instrumental for selecting appropriate approaches to improve the efficiency and effectiveness of fluoroscopic X-ray pose estimation in widespread image-guided interventions.

**1. Introduction**

Fluoroscopic image-guided interventions are increasingly utilized in modern-day healthcare as minimally invasive alternatives to open surgical procedures. For example, endovascular coiling is now the preferred treatment method for ruptured cerebral aneurysms, and endovascular aspiration thrombectomy is an effective intervention for large vessel strokes [1, 2]. Percutaneous biopsies and therapeutic injections (epidural steroid injections, sclerotherapies, etc.) are routine. Although 3D CT, CBCT, or MRI data are readily available for the pre-procedural planning, real-time *intraoperative* guidance is most often limited to 2D X-ray fluoroscopic projections of these 3D volumes. Furthermore, the patient position and orientation may change substantially between consecutive imaging procedures, or even intraoperatively as many procedures are performed with the patient awake. Thus, many tasks in image-guided procedures can be cast as pose estimation problems where a series of 2D projections are used to reach an orientation or target in a 3D volume. This complex pose estimation and 3D/2D registration task, which often has a low tolerance for error, is currently performed by the physician operators. However, machine learning aids for this critical process could greatly improve procedure efficiency and safety [3].

Prior efforts to leverage machine learning for this task largely utilized supervised training frameworks [3]. Due to the sizeable data requirements of this approach, as well as the limited availability of manually labeled medical datasets, many of these works relied on the generation of synthetic digitally reconstructed radiographs (DRRs) from 3D anatomic volumes [4-8]. To enhance the fidelity of DRRs, previous studies [6, 9, 10] have employed domain transfer methods, such as pix2pix and cycle-consistent generative adversarial networks (GANs) [11, 12], aiming to minimize the stylistic differences between DRRs and actual X-ray images. However, despite the ease of implementation of cycle-consistent GAN-based translation techniques, which do not require ground-truth data, their robustness has been found to be limited for X-ray imaging modalities [6].

Recently, there have been rapid advances in differentiable rendering for photorealistic view synthesis and real-world scene representation. Mildenhall et al. [13] have introduced the neural radiance field (NeRF) to represent real-world 3D scenes using a multi-layer perceptron (MLP) and to synthesize high-fidelity RGB images. Using this MLP scene representation, they also implemented a fully vectorized and differentiable rendering algorithm in a machine learning framework (e.g., TensorFlow or PyTorch). Implementations of NeRF have now achieved state-of-the-art performance for the rendering of optically reflective materials [14, 15]. Furthermore, by leveraging the differentiability of NeRF, Lin et al. [16] subsequently introduced a gradient-based optimization framework that performs state-of-the-art 6DOF pose estimation given the NeRF scene representation and a single RGB view of the scene, which they termed "inverting" NeRF (iNeRF).

Motivated by these recent works on optically reflective scene representation, here we introduce methods for pose estimation of radiolucent objects using 2D fluoroscopic projections, and we examine the impact of different 3D scene representations for this task. First, we develop DiffProj, a view synthesis framework for generating DRRs with automatic differentiation. In conjunction with classic CBCTs, we then perform pose estimation by iterative gradient descent using loss functions that quantify the similarity of synthesized DRRs and the true fluoroscopic image of the target pose. We introduce two novel methods for neural scene representation and high-fidelity DRR view synthesis, Neural Tuned Tomography (NeTT) and masked Neural Radiance Fields (mNeRF). Additionally, we also investigate the effects of different loss functions on the performance of iterative pose estimation with all the three scene representations (CBCT, NeTT, and mNeRF).

## 2. Methods

### 2.1. Differentiable Projection (DiffProj) for DRR Synthesis

As illustrated in **Fig. 1**, our model consists of a movable cone-beam X-ray source and image intensifier (such as a fluoroscopy C-arm) with a 3D volume/field representing an anatomic structure fixed at the X-ray isocenter (i.e., the rotation center of the movable X-ray). The X-ray source and image intensifier translate and rotate together relative to the X-ray isocenter. Consistent with the NeRF rending algorithm, we pad and scale our anatomic 3D CBCT volumes, mapping them onto a normalized device coordinate (NDC) system, with *X,Y,Z* values in the $[-1,1]^3$ cube and an origin set at the X-ray isocenter (O) [13]. The source-isocenter distance (SOD) and the source-intensifier distance (SID) are determined at the time of CBCT acquisition and define the geometry of the X-ray source and image intensifier relative to the anatomic volume. The SOD and SID must therefore also be mapped into the NDC so that the resulting projection geometries are maintained (*Appendix A*). We also normalize the scalar densities within the 3D volumes/fields to have values in the range [0, 1].

To synthesize DRRs in this framework, we adopted a Beer-Lambert Ray Casting (RC) algorithm. Consider a single ray ($\boldsymbol{r} = \boldsymbol{o} + t\boldsymbol{d}$) which is emitted from the X-ray source ($\boldsymbol{o}$) along an arbitrary direction ($\boldsymbol{d}$), as shown in **Fig. 1**. According to the Beer-Lamber Law [17, 18], the resulting intensity of a pixel of the DRR along a single ray direction can be determined by the cumulative absorbance of the initially emitted X-ray intensity. RC Absorbance can be simply formulated below:

$$A(\boldsymbol{r}) = \sum_{i=1}^{N_s} \sigma_i \delta_i \qquad (1)$$

where $\sigma_i$ represents sampled volume densities, $N_s$ is the number of samples along the ray, and $\delta_i = t_{i+1} - t_i$ is the spacing between adjacent sampled densities ($t_i$ is the distance of $\sigma_i$ to the source along the ray). The cumulative Absorbance is then related to the X-ray image pixel intensity by $A = -\log(I/I_0)$, where $I_0$ is the intensity of the emitted X-ray beam, and $I$ is the X-ray intensity at the pixel on the image intensifier. A detailed derivation of **Eq. 1** can be found in the previous work [19].

To generate all pixel intensities in a DRR image, **Eq. 1** must be evaluated for a total of *L* × *W* rays, where *L* and *W* are the number of pixels along the length and width of the DRR, respectively. In this work, the DRR image size was set to 128 × 128, computed with a ray sampling size of $N_s$ = 128.

The DiffProj rendering module was coded with full vectorization in TensorFlow to ensure differentiability for subsequent machine learning implementations. In the remainder of this work, we leverage DiffProj to render both discretized 3D volumes and continuous neural density fields, including CBCT, NeTT, and mNeRF (*Sections 2.2 ~ 2.4*). DiffProj also forms the backbone of our framework for gradient-based pose optimization (*Section 2.5*).

*2.2. Tomographic X-ray Acquisition and CBCT Reconstruction*

We analyzed the fluoroscopic X-ray sequences of 50 de-identified noncontrast cerebrovascular CBCT acquisitions, as illustrated in **Fig. 2a**. Each X-ray sequence includes 133 X-ray images with a size of 960 × 960 covering a range of principal (axial) angles from -100° to 100° in an increment of 1.5°. We pre-processed the raw, dynamic-range X-ray images (with an intensity range of [0, 4095]), sequentially performing contrast enhancement (tone mapping and histogram equalization), intensity (grayscale) inversion, intensity scaling to [0, 255], and image resizing to 256 × 256. A sample post-processed X-ray sequence is presented in **Fig. 2b**. For traditional CBCT reconstruction, the processed X-ray sequence was imported into an open-source GPU-based CT reconstruction toolbox, TIGRE [20, 21], which also adopted the same Beer-Lambert Ray Casting (**Eq. 1**). Within TIGRE, the SOD and SID (which were unavailable in the DICOM metadata) were set to 1000 mm and 1536 mm, respectively, and the OSSART reconstruction algorithm was adopted. A sample CBCT reconstruction with a size of 256 × 256 × 256 is presented in **Fig. 2c**.

*2.3. Neural Tuned Tomography (NeTT)*

Discrepancies in style and texture may exist between the real X-rays of the CBCT tomographic sequence and the synthetic DRR projections generated by DiffProj. Here, we describe a deep learning method for tuning the CBCT densities to facilitate domain style transfer from synthetic DRR to real X-ray. We utilize a Multi-Layer Perceptron (MLP) consisting of fully connected layers to tune the CBCT densities sampled in DiffProj, as illustrated in **Fig. 3**. Analogous to the positional encoding utilized in NeRF [13], we encode the CT density ($\sigma$) as a vector for input into the MLP:

$$E(\sigma) = [\sigma \quad \sin 2^i\sigma \quad \cos 2^i\sigma \quad \cdots] \tag{2}$$

where a six-layer encoding is adopted, i.e., $0 \leq i \leq 5$. The MLP then outputs a new scalar density ($\hat{\sigma}$), which is used for the intensity calculation of each ray in DiffProj (**Eq. 1**).

We trained the MLP using the 133 processed X-ray images and the corresponding ground-truth poses in each CBCT acquisition (**Fig. 2b**). The X-ray images were downsized to 128 × 128 to match the size of the DRRs generated by DiffProj. The SOD and SID were scaled to 7.8 and 768.0 in the NDC of DiffProj to maintain the proper geometry (**Appendix A**). During MLP training, an X-ray image ($I_{Xray}$) and the corresponding ground-truth source pose ($p_t$) were randomly sampled from the processed X-ray sequence. Optimization of the MLP weights ($W$) can be formulated as:

$$W_* = \underset{W}{\mathrm{argmin}}\, \mathcal{L}\left(I_{DRR}(p_t|W),\, I_{Xray}(p_t)\right) \tag{3}$$

where $I_{DRR}(p_t|W)$ is the NeTT-optimized synthetic DRR generated by DiffProj at the ground-truth source pose ($p_t$) given the MLP weights ($W$). The pose ($p_t$) is one of the principal angles in an interval of 1.5° from -100° to 100° derived from the CBCT acquisition sequence (**Fig. 2a**). $\mathcal{L}$ is a combined image loss function, quantifying the difference between a pair of $I_{DRR}$ and $I_{Xray}$. It consists of a mean square error (MSE) loss, a focal frequency loss [22], and a structural similarity (SSIM) loss [23]. **Eq. 3** can be efficiently solved with gradient-based optimization, because the differentiability of DiffProj enables computation of $\frac{\partial I_{DRR}}{\partial W}$. During training, we noted that feeding zero-densities (air) in the CT volume to the MLP probably resulted in DRR projections degenerating to a uniform intensity distribution. To avoid this issue, we filtered the sampled CT densities, such that only non-zero CT densities are tuned by the MLP.

### 2.4. Masked Neural Radiance Field (mNeRF)

Differing from finite, discretized CBCT, CT, or MRI volumes, NeRF scene representations are unconstrained and continuous. As illustrated in **Fig. 4a**, NeRF utilizes an MLP to map coordinates (input) to densities (output) [13]. To predict the density at a sampled position, the corresponding coordinates are provided to the MLP after positional encoding [13]:

$$E(x,y,z) = [x \quad y \quad z \quad \sin 2^i x \quad \sin 2^i y \quad \sin 2^i z \quad \cos 2^i x \quad \cos 2^i y \quad \cos 2^i z \quad \cdots] \quad (4)$$

where we adopt a six-layer encoding, i.e., $0 \leq i \leq 5$. To synthesize DRRs (**Eq. 1**), the densities ($\hat{\sigma}$) predicted by the NeRF MLP are sampled along each ray traced by the DiffProj algorithm. The NeRF MLP is trained in a similar manner to NeTT, as described above (**Eq. 3**). Although an MSE loss was used alone to train NeRF in prior works with RGB images [13], we found that incorporation of the SSIM loss aids in convergence for our application.

Initial experiments demonstrated that the unconstrained NeRF generated "floating" artifacts outside of the CBCT reconstructed anatomic structure, as shown in **Fig. 4b**. These artifacts can compensate for the style differences between DRRs and X-rays, but they also reflect overfitting to our limited ground-truth X-ray sample, and do not yield plausible results when rendering new, out-of-sample poses. To eliminate these artifacts, we introduced "masked NeRF" (mNeRF), in which the 3D region of the head, segmented from the CBCT reconstruction, is used as a 3D mask to spatially constrain the non-zero values of NeRF during training (**Fig. 4b**). The 3D mask was processed using 3D Slicer [24] with a slight dilation of 3 mm applied. All densities outside of this mask are forced to zero. An mNeRF masking module was implemented in the TensorFlow machine learning framework to ensure differentiability.

*2.5. Pose Estimation by DRR/X-ray Image Registration*

Intuitively, we can estimate the pose of a single X-ray projection relative to a 3D volume/field by iteratively nudging a randomly initialized pose to minimize the image difference between a rendered DRR and the target X-ray image (**Fig. 5a**). In our model, the pose describes the location/orientation of the X-ray source and has six components, including three angular and three positional degrees of freedom (DOFs), and the anatomic volume is fixed. However, with a simple coordinate transformation, this is equivalent to pose estimation of a movable anatomic structure under a stationary X-ray source. See *Appendix B* for further details on the equivalence of source vs. object movements, and on the coordinate transformation to move between these two configurations.

Given an X-ray image ($I_{Xray}$) with an unknown pose, the estimation of a 6DoF pose ($p$) can be described as an optimization problem:

$$p_* = \underset{p}{\text{argmin}}\, \mathcal{L}(I_{DRR}(p), I_{Xray}) \quad (5)$$

where $I_{DRR}(p)$ is the synthetic DRR generated by DiffProj at a source pose ($p$). $p = [\theta; U]$ consists of three angular DoFs, $\theta = [\theta_y \quad \theta_x \quad \theta_z]$ and three positional DoFs, $U = [U_x \quad U_y \quad U_z]$. Here we define $\theta$ with three *intrinsic* Euler angles applied in a "YXZ" sequence [25]. $\mathcal{L}$ is the image loss function used to quantify the difference between $I_{DRR}$ and $I_{Xray}$. After comparing multiple common loss functions for pose optimization we adopted a mutual information (MI) loss [26, 27], which substantially outperformed the alternatives. Because DiffProj is differentiable, the gradient, $\frac{\partial I_{DRR}}{\partial p}$ can be computed by backpropagation in machine learning frameworks, enabling us to efficiently solve **Eq. 5** by gradient descent.

We initialize the pose ($p_0$) with randomly chosen angular and positional movements away from the target pose ($p_t$):

$$p_0 = p_t + \Delta p \quad (6)$$

where $p_0 = [\theta_0; U_0]$ and $p_t = [\theta_t; U_t]$. $\Delta p$ is a vector consisting of six uniformly random numbers between min($\Delta p$) = [-30, -30, -30; -0.2, -0.2, -0.2] and max($\Delta p$) = [30, 30, 30; 0.2, 0.2, 0.2]. The angular components are measured in degrees, and the positional components reflect a fractional distance within the full NDC space which has bounds of -1 and 1 in each dimension. This pose initialization range was chosen to simulate the starting points that physician operators would encounter during routine image-guided procedures, and the initialization range is large enough that the initial pose is most often substantially different from the target pose, as demonstrated in **Fig. 5b**. We normalized the optimization variables, $p$, onto a regular variable space bounded by $\overline{lb} = 0_{1\times6}$ and $\overline{ub} = 1_{1\times6}$ for efficient optimization [28].

In our work, the angular and positional DoFs were coupled into a single optimization (**Eq. 5**). However, our clinical practice (e.g., percutaneous biopsies and therapeutic injections) demands significantly higher accuracy for angular DoFs compared to positional DoFs, because our goal is to accurately identify the true Frontal and Lateral views of patients through pose estimation. In addition, we, and others, have found that estimation of the angular pose parameters is much more challenging than estimating the positional pose parameters [29]. Therefore, we focused on the errors in angular DoFs, which we measure by introducing a 3D angle error ($\mathcal{E}_\theta$). Given the angular DoFs, $\boldsymbol{\theta}_*$ from the estimated pose and $\boldsymbol{\theta}_t$ from the ground-truth pose, a 3D angle based on the axis-angle representation [30] can be calculated:

$$\mathcal{E}_{\boldsymbol{\theta}} = \arccos \frac{\text{trace}(\boldsymbol{R}_* \cdot \boldsymbol{R}_t^T) - 1}{2} \qquad (7)$$

where $\boldsymbol{R}_*$ and $\boldsymbol{R}_t \in \mathbb{R}_{3\times3}$ are the rotation matrices derived from the Euler angles $\boldsymbol{\theta}_*$ and $\boldsymbol{\theta}_t$, respectively. The superscript $T$ represents the matrix transpose.

## 3. Experiments and Results

### 3.1. Reconstruction of CBCT, NeTT, and mNeRF

Tomographic X-ray sequences of the skulls of 50 patients were collected, with IRB waiver of written informed consent. Using the X-ray sequences, CBCT, NeTT, and mNeRF were reconstructed for the skull of each patient. We first reconstructed CBCTs of all patients using TIGRE, and on our hardware, each CBCT reconstruction took approximately 5 min. Subsequently, we trained both NeTT and mNeRF MLPs, using the Adam optimizer with a learning rate of $5 \times 10^{-4}$. During each epoch, the 133 X-ray images and corresponding ground-truth poses from the tomographic X-ray sequence were randomly shuffled. NeTT and mNeRF MLPs were trained for a variable number of epochs, until a plateau was reached in the peak signal-to-noise ratio calculated for the DRRs with respect to the X-ray images during an epoch. We used a stopping criterion with a patience of 10 epochs and a min delta of $1 \times 10^{-5}$. We found that mNeRF training was substantially more computationally costly than NeTT training; the average training time was 31.2 min for NeTT and 4.7 hours for mNeRF.

We have demonstrated the excellent generalizability of NeTT for CBCT tuning to synthesize high-fidelity DRR on new subjects without additional re-training (***Appendix C***). Therefore, here, only five NeTT MLPs were trained on the X-ray image series of 5 patients' skulls with their CBCTs as scene representations in DiffProj. Then, one of the trained NeTT MLPs of the 5 patients was randomly selected to tune CBCTs of the other 45 patients and to generate their NeTT-projected DRRs. In contrast, in total 50 mNeRF MLPs must be trained for each individual subject.

The resulting DRRs generated by projecting CBCT, NeTT and mNeRF have similar visual appearance (**Fig. 6a**). The high baseline fidelity of the DRRs generated from CBCTs with the Ray Casting algorithm is anticipated, as the CBCT reconstruction algorithms in TIGRE also adopt the Beer-Lambert Law [21], which more accurately reflects the underlying physics of X-ray imaging.

### 3.2. Pose Estimation Using CBCT, NeTT, and mNeRF

Five representative target X-ray images at poses near -90°, -45°, 0°, 45°, and 90° (**Fig. 6b**) were chosen from each tomographic X-ray sequence. We estimated each target pose multiple times from a randomly initialized starting pose (**Eq. 6**) using CBCT, NeTT, and mNeRF, respectively. The Adam optimizer was used for gradient descent optimization of pose estimation, with a learning rate of 0.03 and an exponential decay rate of 0.5 for the first moment estimates. Here, a large learning rate is utilized, so that the gradient descent is encouraged to jump out of local optima. We terminated the pose optimization when a 50-iteration plateau was reached or at a maximum iteration number of 300. After termination, the pose with the best image loss is considered to be the optimal solution. In terms of average computational cost, pose estimation was nearly two times more computationally costly when using the mNeRF (128.4 sec), compared to the CBCT (61.7 sec) and NeTT (85.7 sec).

#### 3.2.1. The Effects of Loss Functions on Pose Estimation Accuracy

Using this optimization protocol, we first compared L1, L2, SSIM, Dice, and MI loss functions in terms of pose estimation performance. We discovered that the MI image loss resulted in the lowest 3D angle errors, and it is markedly superior to the alternatives for all the three scene representations (**Table 1**).

To elucidate of the optimality of loss functions, we selected the frontal skull X-ray of a patient as the ground-truth X-ray, and we investigated the dependencies of different loss functions on each rotational DoF in a range of [-30°, 30°]. As shown in **Fig. 7**, the DoF at 0° represents the global optimum (i.e., the derivative of the loss value with respect to the DoF is equal to zero). It can be observed that all loss functions result in local optima (where the derivatives are also zero) for each rotational DoF, except for the MI loss function. This finding highlights the robustness of the MI loss function, effectively preventing entrapment in local optima.

### *3.2.2. The Effects of Scene Representation on Pose Estimation Accuracy*

With the MI loss for gradient descent optimization of pose estimation, we performed a comprehensive evaluation of pose estimation accuracy on a tomographic X-ray dataset of 50 patients' skulls. For each scene representation method, **Fig. 6c** shows the optimal 3D angle errors for all random initializations in the pose estimation of five representative target tomographic X-ray images of all 50 patients. As summarized in **Table 2**, the overall 3D angle error for direct CBCT scene representation is 2.3 ± 6.7° (90% quantile: 3.0°). Pose estimation with NeTT and mNeRF scene representations had similar overall performance (ANOVA, $p > 0.05$), although mean 3D angle errors were slightly better for NeTT (2.5 ± 8.4°, 90% quantile: 2.9°) than for mNeRF (3.2 ± 9.2°, 90% quantile: 3.4°).

From **Fig. 6c**, we also observe that the accuracy of pose estimation at left/right lateral target X-ray poses (i.e., -90° and 90°) is much worse than that at other target X-ray poses (i.e., -45°, 0°, and 45°). The reason for the pose-dependent accuracy is potentially related to the radiolucent projection of symmetric objects, such as the skull which is symmetric about the sagittal plane. When such an object is projected by X-ray onto a symmetric plane (typically causing symmetric features to overlap in the projection), applying identical negative and positive out-of-plane rotations, respectively, could yield projections that are exactly equivalent. Therefore, a (imperfect) lateral X-ray may correspond to two possible pose solutions, in which out-of-plane rotational DoFs are of equal magnitude but opposite; this phenomenon has also been reported in prior work [6].

## 4. Discussion

In this study, we introduced various loss functions and scene reconstruction methods for pose estimation of radiolucent objects, and investigated their influence in 3D/2D registration on pose estimation accuracy. Based on the Beer-Lambert Ray Casting Law, we first developed DiffProj, a framework for efficient computation of DRRs with automatic differentiation. In conjunction with classic CBCT that was reconstructed by an open-source TIGRE toolbox utilizing the same Beer-Lambert Ray Casting, we performed pose estimation by iterative gradient descent using loss functions that quantify the image discrepancy of the DRR synthesized at an arbitrary pose and the true fluoroscopic image at the target pose. Further, we proposed two novel neural scene representations, NeTT and mNeRF, for high fidelity view synthesis.

We determined that the Mutual Information loss is far superior to alternatives for pose estimation in this framework, as it can effectively prevent entrapment in local optima. Compared to the direct rendering of CBCT, we find that both NeTT and mNeRF result in comparable performance in DRR appearance and pose estimation, although they typically require considerable computational resources and time for neural network training. Overall, in comparison with previously published studies that utilized intensity-based methods, our pose estimation outcomes consistently present state-of-the-art performance for all scene representations in 3D/2D registration, including CBCT, NeTT, or mNeRF (**Table 2**). In particular, we employed a much larger dataset to evaluate the accuracy of single-view fluoroscopic pose estimation.

This study reveals that when a physically accurate X-ray rendering algorithm (e.g., Beer-Lambert Ray Casting) is consistently adopted in both scene reconstruction and projection, the accuracy of pose estimation is less affected by the scene representation, whether through discretized grids or neural networks. However, it is worth noting that the reconstruction methods used in commercially available CT scanners typically serve as a black box, due to the proprietary processing of raw, dynamic-range X-ray images (as performed in **Fig. 2b**) and potentially other modalities (e.g., CT, PET-CT etc.). In these scenarios, direct rendering of the reconstructed CT often yields DRRs that substantially differ from ground-truth X-ray images, and it is essential to improve DRR fidelity for fluoroscopic X-ray pose estimation [6, 9, 10].

In *Appendix C*, we simulated a scenario by replacing the Ray Casting algorithm with the Volume Rendering algorithm [13, 31] in DiffProj, assuming that we are unaware of the use of Ray Casting in CBCT reconstruction. Notably, this alternative Volume Rendering method is based on a different physical principle than that of the Beer-Lambert Ray Casting model, but it has been also implemented in rendering realistic fluoroscopic DRRs [32]. The modification results in DRRs, through the direct projection of CBCT, having a substantial style discrepancy from our post-processed ground-truth X-ray images (**Fig. 2b**). We showed that both NeTT and mNeRF are capable of substantially improving the fidelity of DRRs by Volume Rendering, and hence, significantly boosting the accuracy of pose estimation.

Moreover, regardless of whether Ray Casting or Volume Rendering is used in DiffProj, NeTT training requires considerably less computational recourses than mNeRF. Crucially, NeTT demonstrates excellent cross-subject generalizability, which eliminates the need for NeTT MLPs to be trained on each individual subject, as required by mNeRF MLPs. Given the much lower computational cost for NeTT in both training and pose optimization, we suggest that NeTT is a more attractive option for domain alignment between DRR and X-ray than mNeRF.

Our image dataset consisted of tomographic X-ray series from 50 patients' CBCT acquisitions, which are routinely collected in medical practice. However, a limitation of this dataset arises from the fact that all of the ground truth tomographic images are collected on a single semi-circular arc around the patient. Consequently, real X-ray images from poses outside of this arc were not available for evaluation in this study. In the evaluation of pose estimation, the five representative target poses selected from within the tomographic X-ray series do represent common projections utilized during neurointerventional procedures (i.e., Anteroposterior, Lateral, and Oblique views), and large angles in the Craniocaudal or Caudocranial directions are less commonly used in routine clinical practice.

In addition, we performed CBCT reconstruction ($256 \times 256 \times 256$) and view synthesis ($128 \times 128$) using resolutions that are below what is typically encountered in clinical scenarios. For example, our fluoroscopic images often have a size of $1024 \times 1024$. We down-sampled images in this study due to the memory limitations of our hardware. Future optimizations may enable view synthesis and pose optimization with higher resolution images, which we anticipate will improve the pose estimates.

**Competing Interests**

Portions of the work described in this article have been included in a related patent filed by Northwestern University, with C. Zhou, D.R. Cantrell, and S.A. Ansari listed as co-inventors. D.R. Cantrell and S.A. Ansari are founders and have shares in Clearvoya, LLC, which aims to commercialize Computer Vision algorithms for Image-Guided interventions, but this work was not funded or performed by Clearvoya. M.C. Hurley made contributions to this work while he was at Northwestern University, but he did not contribute from his new institution.

**Ethical Statement**

CBCT acquisitions routinely collected during neurointerventional angiography were utilized in this study. This study was reviewed by our IRB (#: STU00212923), which determined that informed consent could be waived.

**Acknowledgement**

None.

29. Hanley J, Mageras GS, Sun J, Kutcher GJ: The effects of out-of-plane rotations on two dimensional portal image registration in conformal radiotherapy of the prostate. Int J Radiat Oncol Biol Phys 33:1331-1343, 1995

30. Mahendran S, Ali H, Vidal R: 3D Pose Regression Using Convolutional Neural Networks, 2017

31. Tagliasacchi A, Mildenhall B: Volume Rendering Digest (for NeRF), 2022

32. Corona-Figueroa A, et al.: MedNeRF: Medical Neural Radiance Fields for Reconstructing 3D-aware CT-Projections from a Single X-ray: Institute of Electrical and Electronics Engineers Inc., 2022

**Tables**

**Table 1**: Comparison of 3D angle errors (°) of pose estimations using different image loss functions. Here, scene representations of 5 patients' skulls were included for registration to 5 representative target X-ray images; 20 registrations with random initialization were performed for each target X-ray image of each patient. (Note: MSE = mean squared error; SSIM = structural similarity; MI = mutual information)

|  | CBCT | NeTT | mNeRF |
|---|---|---|---|
| SSIM | 11.1 ± 14.9 | 11.0 ± 15.2 | 15.3 ± 15.6 |
| Soft Dice | 8.0 ± 9.1 | 9.4 ± 10.6 | 12.4 ± 11.0 |
| L1 | 7.6 ± 8.9 | 7.8 ± 9.7 | 11.8 ± 11.5 |
| MSE | 6.1 ± 8.6 | 6.3 ± 9.2 | 9.1 ± 9.7 |
| **MI** | **2.6 ± 7.5** | **1.8 ± 6.1** | **3.0 ± 8.6** |

**Table 2**: Comparison of angular errors of previously reported works with our pose estimation outcomes.

|  | Imaging Setup | Objects | Specimens | Methods | Performance |
|---|---|---|---|---|---|
| Grupp et al [7] | Single Fluoroscope | Pelvises | 6 | Hybrid [a] | < 1° for 86% of the images |
| Gao et al [8] | Single Fluoroscope | Pelvises | 10 | Intensity-based | Mean: 6.94 ± 7.47°; Median: 3.76° |
| Lin et al [16] | Single Camera | Real-world scenes | 4 | Intensity-based | Mean: 4.39°; Median: 2.01° |
| Zhou et al [6] | Dual Fluoroscopes | Skull | 1 | Landmark-based | Mean: 3.9 ± 2.1° [b] |
| **Current study** | **Single Fluoroscope** | **Skulls** | **50** | **Intensity-based** | **CBCT:** 2.3 ± 6.7°; 3.0° (90% quantile)<br>**NeTT**: 2.5 ± 8.4°; 2.9° (90% quantile)<br>**mNeRF**: 3.2 + 9.2°; 3.4° (90% quantile) |

Note:

a. The hybrid method combined both the landmark-based and intensity-based methods.

b. The error is defined as the Euclidian distance of vectors comprising three extrinsic Euler angles.

**Figures**

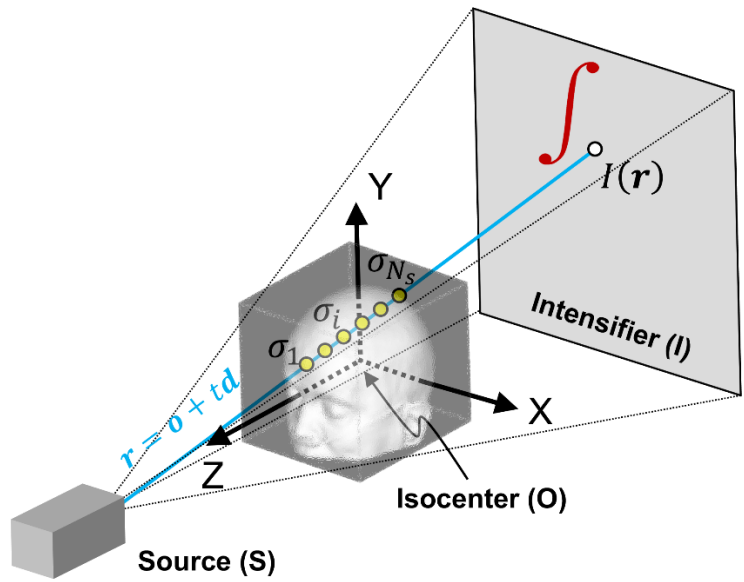

**Fig. 1**: Illustration of the DiffProj geometry, consisting of an X-ray source (S) and an image intensifier (I). The NDC coordinate system origin is set at the isocenter (O). For an arbitrary ray ($r$) cast from the source onto a single pixel of the image intensifier, the pixel intensity ($I$) is determined by the sampled densities ($\sigma_i$, $i = 1, 2, \ldots, N_s$) of the 3D volume.

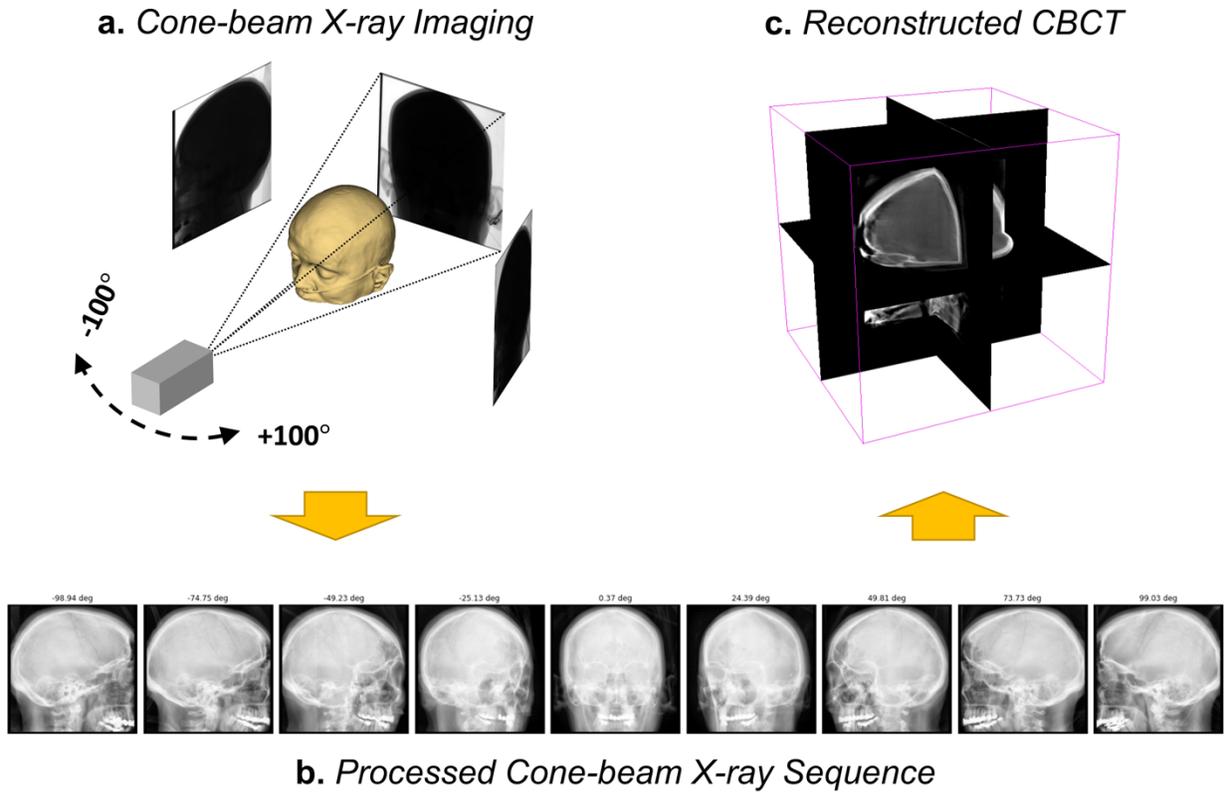

**Fig. 2**: Our data acquisition and pre-processing pipeline includes: **a**) Acquisition of raw, dynamic-range tomographic X-ray series with a size of 960 × 960 over a range of -100° to 100°; **b**) Post-processed X-ray images sequentially undergoing contrast enhancement, grayscale inversion, intensity scaling, and downsampling to a size of 256 × 256; and **c**) CBCT with a size of 256 × 256 × 256 reconstructed by TIGRE.

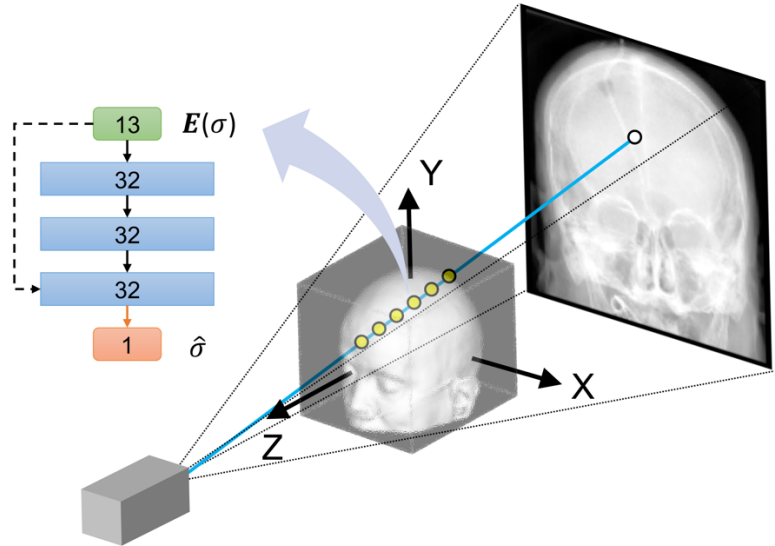

**Fig. 3**: Illustration of our NeTT optimization with DiffProj. A sampled density ($\sigma$) along a ray is encoded and input into an MLP (flowchart) to output a tuned density ($\hat{\sigma}$), i.e., $\hat{\sigma} = \text{MLP}(\sigma|W)$. The green block represents the input density encoding, the blue blocks are hidden layers, and the orange block is the output density. The numbers in the blocks indicate the tensor sizes. The black arrows denote Dense + ReLU + Batch Normalization layers, the dashed arrow denotes a skip connection (i.e., concatenation with the input density encoding), and the orange arrow denotes Dense + ReLU layers.

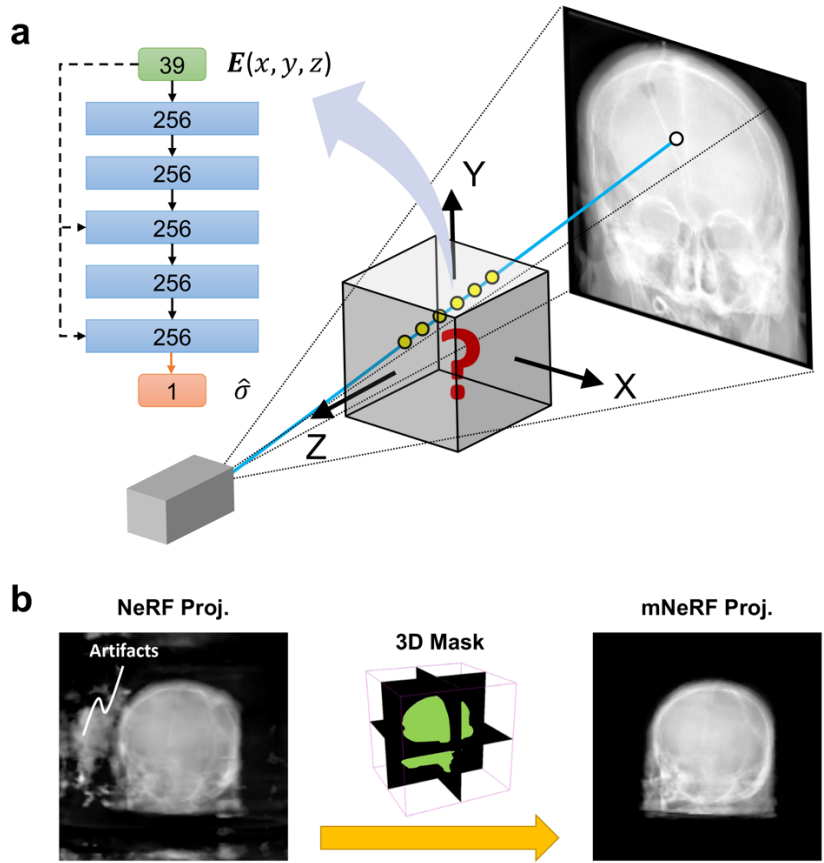

**Fig. 4**: **a**) Illustration of synthetic DRR generation with DiffProj from a NeRF scene representation. The coordinates $(x, y, z)$ at a sampled position are used as input to the NeRF MLP (shown by the flowchart) to predict the density ($\hat{\sigma}$) at the sampled position, i.e., $\hat{\sigma} = \text{MLP}(x, y, z \mid W)$. The green block represents the input positional encoding, the blue blocks are hidden layers, and the orange block is the output density. The numbers in the blocks indicate the tensor size. The black arrows denote Dense + ReLU + Batch Normalization layers, the dashed arrows denote skip connections (i.e., concatenation with the input positional encoding), and the orange arrow denotes Dense + ReLU layers. **b**) Skull projections of NeRF (without a mask) and mNeRF scene representations after training, at an out-of-sample pose not seen during training. Both images have a size of 128 × 128 and normalized intensities with a range of [0, 1]. The application of a differentiable, spatially constrained mask results in substantial artifact reduction. (Note: Proj. = Projection)

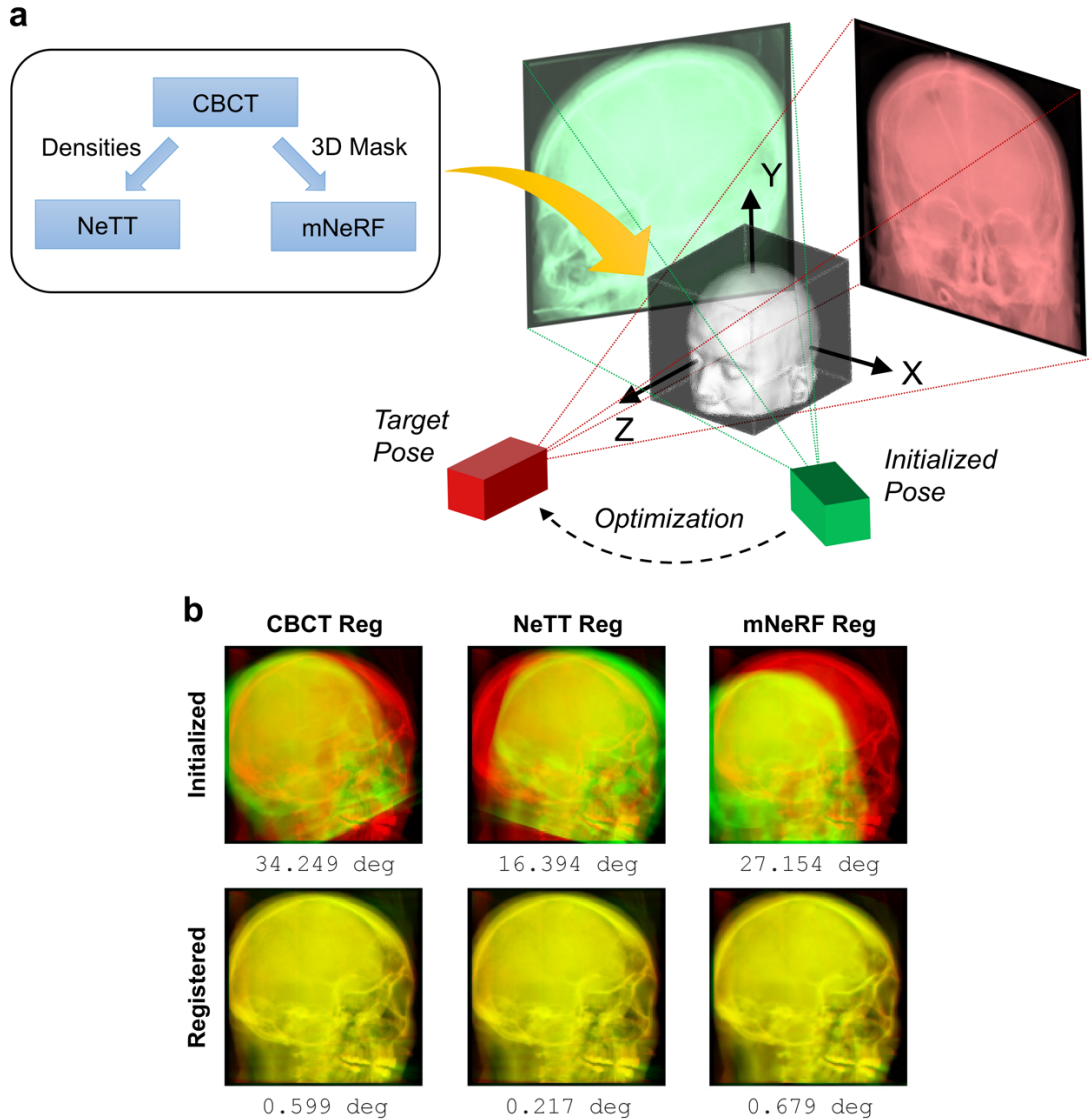

**Fig. 5**: **a**) Illustration of single-view X-ray pose estimation with DiffProj. The target X-ray pose can be reached by iteratively nudging a randomly initialized DRR pose to minimize the difference between the X-ray and the rendered DRR. As a radiolucent scene representation for DiffProj computation, CBCT, NeTT, and mNeRF can be used. Both innovative techniques, NeTT and mNeRF, rely on classic CBCT reconstructions. As reflected in the *inset,* NeTT directly optimizes the CBCT densities, while the non-zero values of mNeRF are constrained by a 3D mask of the anatomic region segmented from CBCT. **b**) Examples of accurate registration of CBCT, NeTT, and mNeRF projections (*green*) onto a target X-ray image (*red*) using DiffProj. Overlapping regions render in *yellow*. The 3D angle errors calculated between the poses of DRR and X-ray are listed below each plot. The *upper* row illustrates the random initial poses, and the *lower* row illustrates the optimal pose estimation. (Note: Reg = Registration)

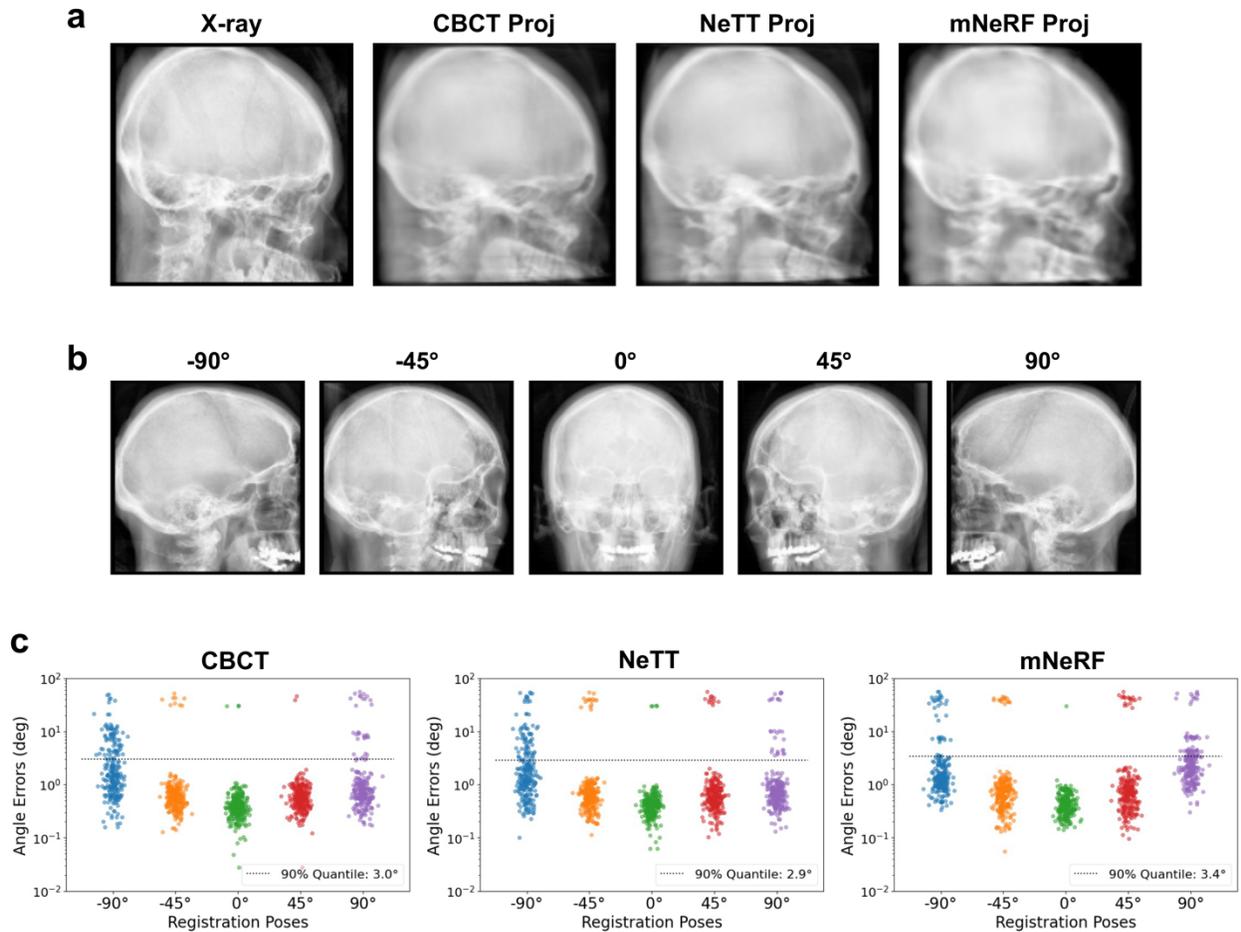

**Fig. 6**: **a**) Comparison of a ground-truth X-ray image with DRRs generated at the same pose by projecting CBCT, NeTT, and mNeRF, respectively, using Ray Casting. All images have a size of 128 × 128 and normalized intensities with a range of [0, 1] (Note: Proj = Projection). **b**) Selected X-ray images at five target poses used to evaluate the performance of pose estimation using our gradient-based optimization framework. **c**) Comparison of the 3D angle errors using Ray Casting to project CBCT, NeTT, and mNeRF scene representations, respectively, for all pooled trials of all 50 patients; for each representative target X-ray image of each patient, 5 registrations with random initialization were performed.

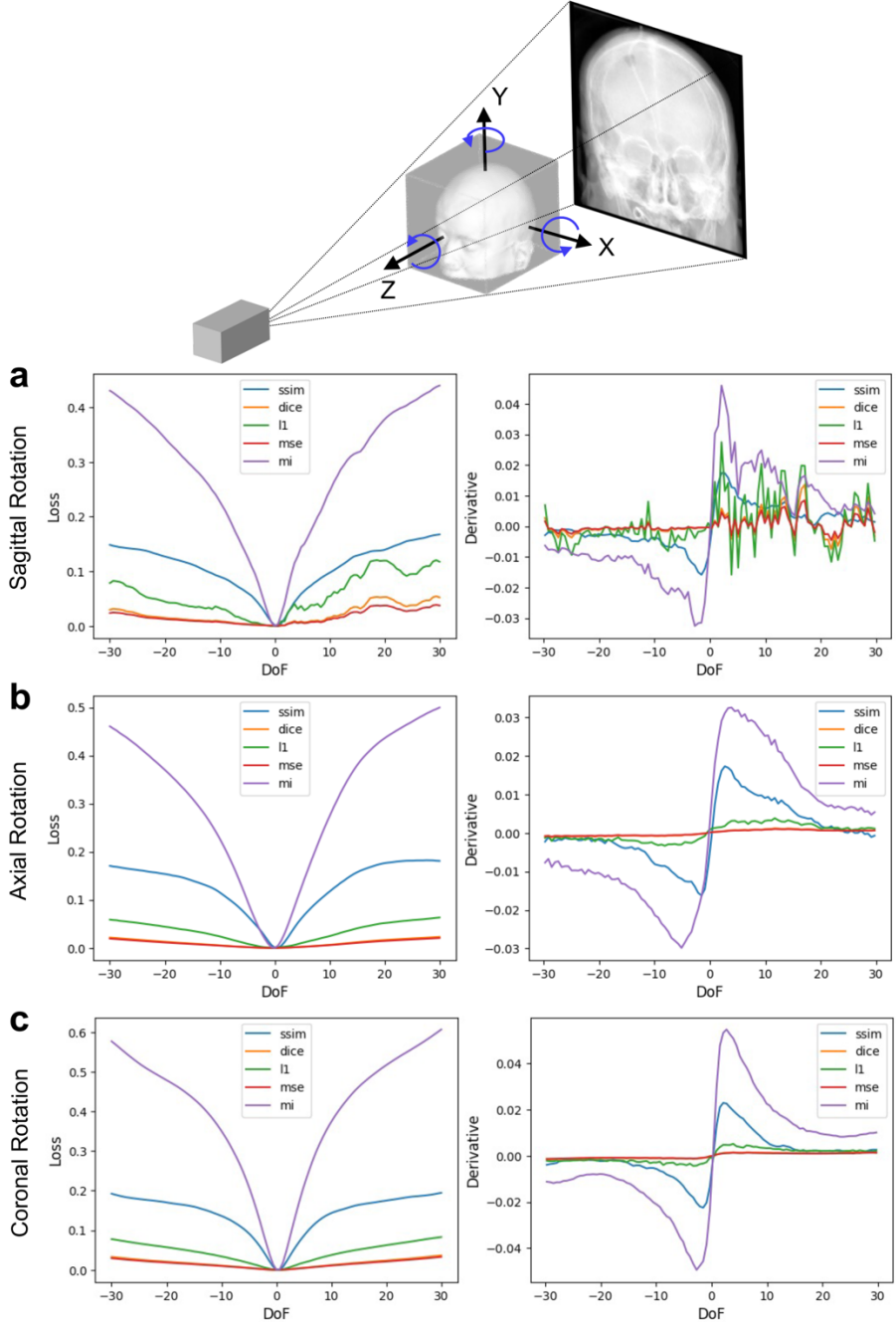

**Fig. 7**: The dependencies of loss functions (*left*) and their derivatives (*right*) on each rotational DoF in a range of [-30°, 30°]. Here, we select the frontal X-ray of a patient as the ground-truth X-ray, corresponding to the DoF at 0° (the global optimum). The losses are calculated for DRRs generated by projecting CBCT at different DoFs with respect to the ground-truth X-ray using different loss functions. For comparison, all losses have been shifted, such that their values at 0° are zero. **a**) Sagittal (X-axis) rotational DoF; **b**) Axial (Y-axis) rotational DoF; **c**) Coronal (Z-axis) rotational DoF.

**Supplementary Material**

**Appendix A. X-ray Configuration Parameters**

In a typical X-ray configuration (e.g., C-arm fluoroscopy), the X-ray source moves relative to a coordinate system (CS) established at the isocenter (i.e., the rotational center of the X-ray), as illustrated **Fig. S1**. The geometry imaged on the intensifier depends on two parameters, the source-isocenter distance (SOD) and the source-intensifier distance (SID; also called the focal distance for a camera). The SOD and SID can be set according to the volume and image sizes, respectively.

For a cubic volume that is scaled to different sizes, $l_1$ and $l_2$ (e.g., those in the physical coordinate and NDC, respectively), the corresponding SODs, $z_1$ and $z_2$, should be set, such that the same geometry is imaged:

$$\frac{z_1}{z_2} = \frac{l_1}{l_2} \tag{S1}$$

Furthermore, the volume may be projected onto a square intensifier using different image sizes, $s_1$ and $s_2$. To ensure that the imaged geometry remains invariant, the corresponding SIDs, $f_1$ and $f_2$, should satisfy:

$$\frac{f_1}{f_2} = \frac{s_1}{s_2} \tag{S2}$$

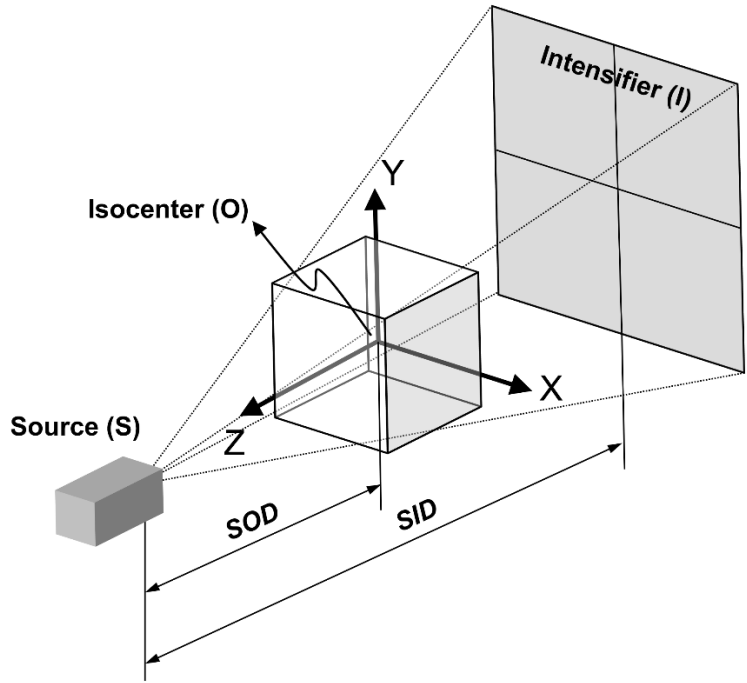

**Fig. S1**: The typical configuration of a movable X-ray imaging system. The imaged geometry on the intensifier is determined by two parameters, the SOD and SID.

**Appendix B. Source vs. Object Movements**

In a movable X-ray configuration, the X-ray source can be moved *first* by translations and *then* by rotations with respect to a coordinate system (CS) established at the isocenter, as shown in **Fig. S2**. In particular, the angular DoFs are represented by three *intrinsic* Euler angles with a sequence of "YXZ". Sometimes, it may also be desirable to consider a moving object relative to a stationary X-ray source (e.g., when aligning a patient's CT to a patient's intraoperative pose in physical 3D space), as shown in **Fig. S3**.

The 6DoF motion (*first* translations and *then* rotations) in the *movable* X-ray configuration is convertible to motion in the *fixed* X-ray configuration (*first* rotations and *then* translations). To ensure the same magnitude of the 6 parameters in both X-ray configurations, under the assumption of a fixed X-ray configuration (**Fig. S3**), the positional CS origin must be set at the X-ray source, and the angular CS origin must be set at the isocenter. In the *fixed* X-ray configuration, both the CSs have positive coordinate directions opposite to those in the *movable* X-ray configuration (i.e., the two CSs in the *fixed* X-ray configuration are left-handed). In the *fixed* X-ray configuration, the angular DoFs of the object are defined by three *extrinsic* Euler angles with the same sequence of "yxz" with respect to the angular CS, while the positional DoFs of the object are measured with respect to the positional CS.

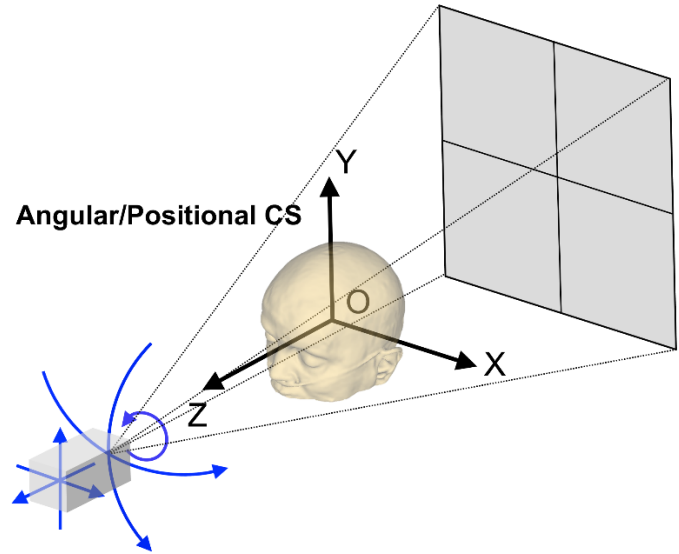

**Fig. S2**: The configuration of a movable X-ray, where a source moves relative to a fixed object. The positional/angular CS is set at the isocenter. The *blue* arrows indicate the positive directions of 6DoFs.

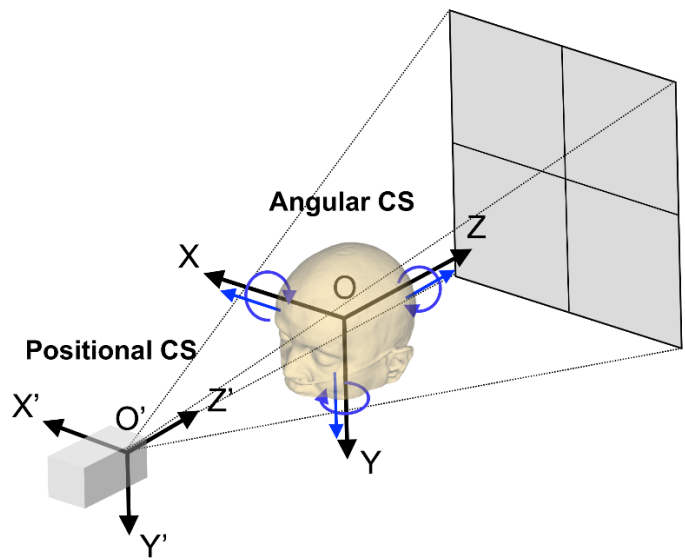

**Fig. S3**: The configuration of a fixed X-ray, where an object moves relative to a fixed source. The positional CS origin is set at the source, while the angular CS origin is set at the isocenter. The *blue* arrows indicate the positive directions of 6DoFs.

**Appendix C. Pose Estimation with Volume Rendering Algorithm**

*C.1. Differentiable Projection (DiffProj) with the Volume Rendering Algorithm*

As demonstrated in **Fig. 6a**, direct projection of CBCT using the Beer-Lambert Ray Casting algorithm, resulted in DRRs with high fidelity. To further evaluate the ability of NeTT / mNeRF to perform domain style transfer and to refine DRR synthesis, we generated DRRs using an alternative rendering algorithm that results in a significant style difference of the DRRs when compared to the ground-truth X-ray images. For this purpose, we adopted the Volume Rendering algorithm described by Mildenhall et al in their original paper on NeRF [1], which we modified for X-ray (grayscale) rendering. Notably, this alternative Volume Rendering method, which more appropriately describes optically reflective materials, differs from the Beer-Lambert model, which more accurately describes the underlying X-ray physics [2, 3]. When adopting the NeRF Volume Rendering Equation, pixel "Intensity" should be interpreted only as the pixel luminance or brightness, as it does not directly correspond to the physical X-ray intensity.

Within in the DiffProj framework (**Fig. 1**), consider a single ray ($r = o + td$) which is emitted from the X-ray source ($o$) along an arbitrary direction ($d$). The resulting image intensity ($I$) at the pixel on the image intensifier where the ray arrives can be formulated as a numerical integration of all intensities throughout this ray in the 3D space:

$$I(r) = \sum_{i=1}^{N_s} w_i \sigma_i \qquad (S1)$$

where $w_i$ are weights on each sampled volume density, $\sigma_i$, and $N_s$ is the sample size per ray. To reduce computational cost, the intensities ($\sigma_i$, $i$ = 1, 2, …, $N_s$) can be sampled only within the 3D volume by setting near and far bounds to correspond to the faces of the 3D volume along the Z-axis, as shown in **Fig. 1**.

The weights on each sampled volume density in **Eq. S1** can then be expressed as $w_i = \alpha_i T_i$, comprising an opacity of the sampled volume, $\alpha_i$, and a transmittance, $T_i$, which describes the cumulative absorption of the ray by densities located between the source and the sampled volume with a nested numerical integration:

$$\alpha_i = 1 - \exp(-\sigma_i \delta_i) \tag{S2}$$

$$T_i = \exp\left(-\sum_{j=1}^{i-1} \sigma_j \delta_j\right) \tag{S3}$$

where $\delta_i = t_{i+1} - t_i$ is the spacing between adjacent sampled densities ($t_i$ is the distance of $\sigma_i$ to the source along the ray). The derivation of these equations and physical assumptions are well described by Tagliasacchi and Mildenhall [4]. However, it is worth noting that the volume shading (RGB colors) in their formulations has been replaced with the volume luminance (densities) in our work.

## C.2. Performance of Pose Estimation with Volume Rendering

Using DiffProj with the Volume Rendering algorithm, the resulting visual appearance of DRRs generated by projecting NeTT and mNeRF are similar, but both are significantly better than that generated by projecting CBCT (**Fig. S4**). In contrast, DRRs generated from CBCTs with the Ray Casting algorithm (**Fig. 6a**) are of substantially greater fidelity than those generated with Volume Rendering. This result is anticipated, as the CBCT reconstruction algorithms in TIGRE also adopt the Ray Casting algorithm based on the Beer-Lambert Law [5], which more accurately reflects the underlying physics of X-ray imaging. The suboptimal appearance of the DRR projections generated by Volume Rendering can be substantially compensated by either NeTT or mNeRF, thereby substantially improving the quality of the synthetic images as required by pose estimation tasks (**Fig. S4**).

Further, we also examined the generalizability of NeTT for improving DRR fidelity and pose estimations on new subjects without additional re-training. For this purpose, a NeTT MLP was trained only once using the series of 133 X-ray images in the CBCT of patient #1. The trained NeTT MLP of patient #1 was then utilized for DRR view synthesis from the other 4 patients (#2 ~ #5). As demonstrated in **Fig. S5**, the NeTT MLP of a single subject can promote domain alignment of DRRs with X-ray images for all of other subjects. In contrast, mNeRF must be trained for each individual subject.

To evaluate the performance of pose estimation with Volume Rendering, five representative target X-ray images at poses near -90°, -45°, 0°, 45°, and 90° (**Fig. 6b**) were chosen from each tomographic X-ray sequence of five patient skulls. We estimated each target pose 20 times from a randomly initialized starting pose using CBCT, NeTT, and mNeRF. For each method of scene representation, **Fig. S6** shows the optimal 3D angle errors for all 500 random initializations in the estimation of five target poses on all five patients using the MI loss. It can be observed that NeTT and mNeRF scene representations result in overall 3D angle errors $1.8 \pm 5.9°$ and $1.8 \pm 5.8°$ (90% quantiles: 1.7° and 1.8°), respectively, which are a significant improvement (ANOVA, $p < 0.01$) over the 3D angle error of $6.9 \pm 12.8°$ (90% quantile: 31.1°) achieved using DRRs rendered directly from CBCT (**Table S1**). In contrast, the overall 3D angle error for Ray Casting of direct CBCT scene representation is much better (**Fig. 6c**), since the appearance of CBCT-projected DRRs closely matches that of the X-ray images (**Fig. 6a**).

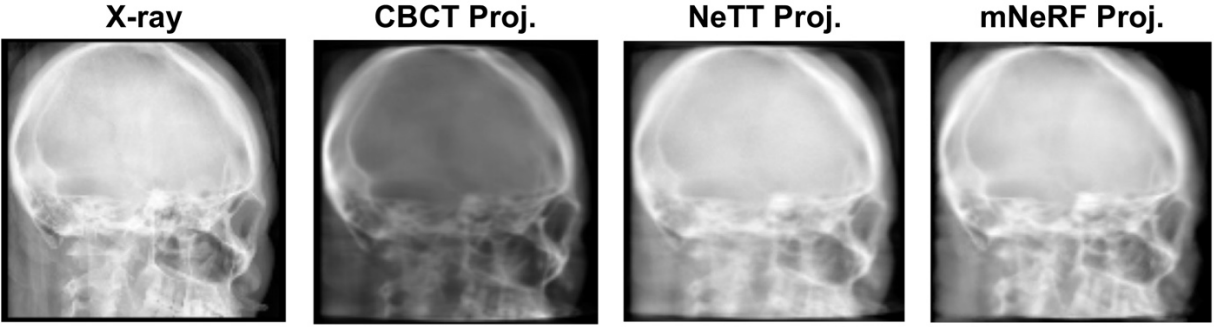

**Fig. S4**: Visual comparison of an X-ray image with DRRs generated at the same pose by projecting CBCT, NeTT, and mNeRF using Volume Rendering. All images have a size of 128 × 128 and normalized intensities with a range of [0, 1].

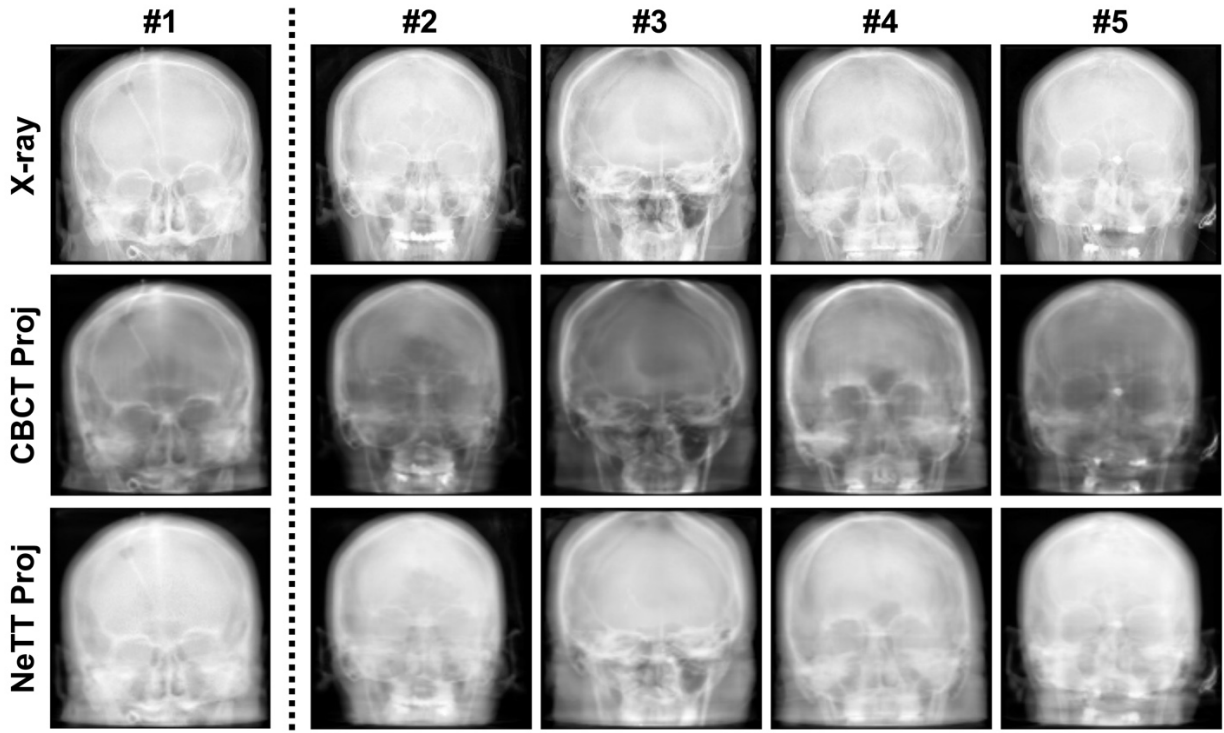

**Fig. S5**: Cross-subject generalizability of NeTT for generating DRRs using Volume Rendering to match the appearance of X-ray images. All images have a size of 128 × 128 and normalized intensities with a range of [0, 1]. Here, a single NeTT MLP was trained using only the 133 X-ray images of patient #1. Then, this single, trained NeTT MLP was used to tune the CBCTs of the other patients (#2 ~ #5) to generate high fidelity DRRs.

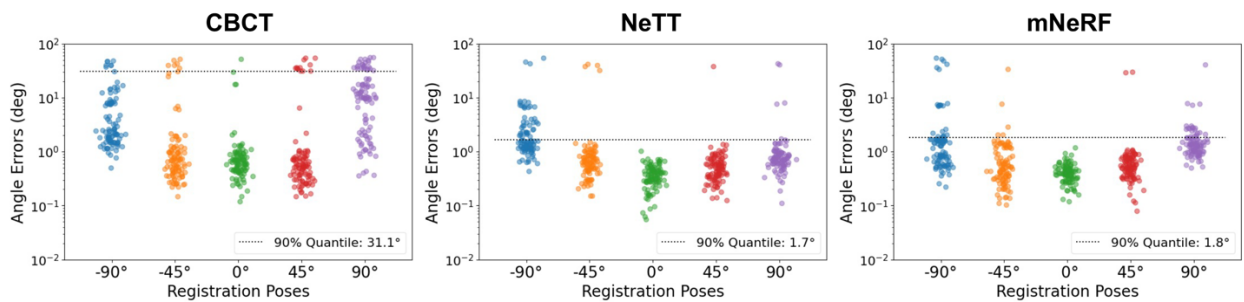

**Fig. S6**: Comparison of the 3D angle errors using Volume Rendering to project CBCT, NeTT, and mNeRF scene representations, respectively. Here, scene representations of 5 patients' skulls were included for registration to 5 representative target X-ray images; 20 registrations with random initialization were performed for each target X-ray image of each patient.

**Table S1**: Performance of Pose Estimation using Volume Rendering to project CBCT, NeTT, and mNeRF, respectively.

|  | CBCT | NeTT | mNeRF |
|---|---|---|---|
| 3D Angle Errors (°) | 6.9 ± 12.8 | 1.8 ± 5.9 | 1.8 ± 5.8 |
| 90% Quantiles (°) | 31.1 | 1.7 | 1.8 |